# Forest canopy height estimation from satellite RGB imagery using large-scale airborne LiDAR–derived training data and monocular depth estimation

Yongkang Lai [a, b], Xihan Mu [a, b], Dasheng Fan [a, b], Donghui Xie [a, b], Shanxin Guo [c], Wenli Huang [d], Tianjie Zhao [e], Guangjian Yan [a, b] *

[a] State Key Laboratory of Remote Sensing and Digital Earth, Faculty of Geographical Science, Beijing Normal University, Beijing 100875, China

[b] Beijing Engineering Research Center for Global Land Remote Sensing Products, Faculty of Geographical Science, Beijing Normal University, Beijing 100875, China [c] CSIRO Environment, Canberra, Australian Capital Territory 2601, Australia

[c] Center for Geo-Spatial Information, Shenzhen Institutes of Advanced Technology, Chinese Academy of Sciences, Shenzhen, 518055, PR China

[d] School of Resource and Environmental Science, Wuhan University, Wuhan 430079, China

[e] State Key Laboratory of Remote Sensing Science, Aerospace Information Research Institute, Chinese Academy of Sciences, Beijing 100101, China

* Corresponding Authors

E-mail: gjyan@bnu.edu.cn

## Abstract

Large-scale, high-resolution forest canopy height mapping plays a crucial role in understanding regional and global carbon and water cycles. Spaceborne LiDAR missions, including the Ice, Cloud, and Land Elevation Satellite-2 (ICESat-2) and the Global Ecosystem Dynamics Investigation (GEDI), provide global observations of forest structure but are spatially sparse and subject to inherent uncertainties. In contrast, near-surface LiDAR platforms, such as airborne and unmanned aerial vehicle (UAV) LiDAR systems, offer much finer measurements of forest canopy structure, and a growing number of countries have made these datasets openly available. In this study, a state-of-the-art monocular depth estimation model, Depth Anything V2, was trained using approximately 16,000 km² of canopy height models (CHMs) derived from publicly available airborne LiDAR point clouds and related products across multiple countries, together with 3 m resolution PlanetScope and airborne RGB imagery. The trained model, referred to as Depth2CHM, enables the estimation of spatially continuous CHMs directly from PlanetScope RGB imagery. Independent validation was conducted at sites in China (approximately 1 km²) and the United States (approximately 116 km²). The results showed that Depth2CHM could accurately estimate canopy height, with biases of 0.59 m and 0.41 m and root mean square errors (RMSEs) of 2.54 m and 5.75 m for these two sites, respectively. Compared with an existing global meter-resolution CHM product, the mean absolute error is reduced by approximately 1.5 m and the RMSE by approximately 2 m. These results demonstrated that monocular depth estimation networks trained with large-scale airborne LiDAR–derived canopy height data provide a promising and scalable pathway for high-resolution, spatially continuous forest canopy height estimation from satellite RGB imagery.

## 1. Introduction

Forest ecosystems are a fundamental component of terrestrial ecosystems and play a critical role in global and regional climate regulation (Bonan 2008; Canadell and Raupach 2008), hydrological processes (Ellison et al. 2017; Hou et al. 2023), and carbon cycling (Augusto and Boča 2022; Pan et al. 2011). Forest canopy height is a key structural parameter that underpins the estimation of aboveground biomass (Asner et al. 2010), reflects forest growth conditions (Poorter et al. 2015; Pretzsch 2014), and influences species richness (Goetz et al. 2007; MacArthur and MacArthur 1961). High-resolution canopy height information can directly characterize habitat heterogeneity (Tuanmu and Jetz 2015), which has been identified as an essential biodiversity variable requiring priority monitoring through spatial observations (Skidmore et al. 2021). Consequently, producing fine-resolution canopy height models (CHMs) over large spatial extents is of substantial importance.

Traditional approaches for measuring canopy height, such as the use of graduated ropes or rangefinders, are labor-intensive and inefficient, making them suitable only for small-scale field experiments rather than spatially explicit mapping. The rapid development of LiDAR systems deployed on multiple platforms has greatly expanded the possibilities for canopy height mapping (Liang et al. 2016; Maltamo et al. 2014). Terrestrial laser scanning (TLS) and UAV-borne LiDAR can capture highly detailed three-dimensional point clouds at the plot scale, enabling the generation of canopy height maps (CHMs) at centimeter-level resolution. However, these approaches remain inefficient and are generally limited to plot-scale applications (*e.g.*, tens to hundreds of meters). Compared with UAV platforms, airborne LiDAR operates at higher altitudes and covers much larger areas, allowing the production of sub-meter resolution CHMs at the stand to regional scale (Khosravipour et al. 2014). Nevertheless, the high operational costs associated with airborne LiDAR surveys make such data inaccessible to many researchers. Spaceborne LiDAR missions comprise both full-waveform systems, such as the Ice, Cloud, and Land Elevation Satellite (ICESat) and the Global Ecosystem Dynamics Investigation (GEDI), and photon-counting systems, such as ICESat-2, providing globally distributed measurements of forest canopy structure (Dubayah et al. 2020; Huang et al. 2023; Lang et al. 2022; Mulverhill et al. 2022; Neuenschwander et al. 2020; Schutz et al. 2005). However, these sensors deliver discrete footprint-level samples rather than spatially continuous observations, which limits their direct applicability for wall-to-wall canopy height mapping.

To address these limitations, numerous studies have combined GEDI relative height (RH) metrics with spatially continuous passive optical imagery using machine learning or deep learning techniques to generate large-scale CHM products. For example, Potapov et al. (2021) employed a machine learning approach (*i.e.*, a bagged regression trees ensemble) to model the relationship between Landsat imagery and GEDI RH95 metrics, producing a global canopy height product at 30 m resolution. Lang et al. (2023) trained ensembles of convolutional neural networks (CNNs) with identical architectures but different initial weights using Sentinel-2 imagery and GEDI RH98 metrics, generating a global CHM at 10 m resolution. Tolan et al. (2024) leveraged Maxar Vivid2 mosaic imagery and a self-supervised Vision Transformer to predict canopy height at 1 m resolution, followed by scale calibration using a CNN trained with GEDI RH95 metrics.

Although these studies produced global products at various spatial resolutions and achieved promising performance, they share several common limitations. From a modeling perspective, most approaches rely on traditional regression models or generic visual representation frameworks, which have limited capacity to explicitly capture three-dimensional forest structure. From a data perspective, the GEDI-derived RH metrics used for training contain inherent uncertainties, including retrieval errors and horizontal geolocation inaccuracies (Jiang et al. 2025), which may be further amplified when applied to high-resolution optical imagery.

In recent years, monocular depth estimation (MDE) has advanced rapidly in the computer vision community. Such deep neural networks estimate the depth of each pixel relative to the imaging platform from a single RGB image. Recently, several studies have explored the application of MDE models to canopy height estimation. Cambrin et al. (2024) trained the state-of-the-art MDE model Depth Anything V2 using airborne RGB imagery and corresponding CHMs from the National Ecological Observatory Network (NEON) sites in the United States, demonstrating that the model can accurately estimate canopy height after task-specific training. Tan et al. (2025) trained a network based on a large vision foundation model using airborne CHM data collected in China and co-located Google Earth imagery, aiming to generate high-resolution CHMs for plantation forests. While these studies achieved high accuracy at local scales or within specific vegetation types and confirmed the feasibility of applying MDE networks to canopy height inversion, they are generally restricted to regional extents or specific scenarios. The performance and generalization capability of deep neural networks strongly depend on the scale and diversity of training data (Zhang et al. 2016a). Models trained on local datasets may face challenges when transferred to other regions or applied at larger spatial scales, thereby limiting their broader applicability.

To address these challenges, this study adopted Depth Anything V2 as the core model. The original task of this network was to estimate the depth of each pixel in an RGB image relative to the camera, which was conceptually well aligned with canopy height estimation from remote sensing imagery. Compared with the regression trees and CNN-based approaches used in previous studies, this model was better suited to extract three-dimensional structural features from images for canopy height estimation. In terms of training data, we compiled a large and diverse dataset consisting of publicly available airborne and UAV LiDAR point clouds from multiple countries worldwide, together with corresponding DSM and DEM products or CHMs derived from airborne LiDAR. In total, approximately 16,000 km² of data were collected for model training.

The remainder of this paper is organized as follows. Section 2 describes the datasets used in this study and the data preprocessing workflow. Section 3 introduces the model architecture, training strategy, and parameter settings. Section 4 first analyzes the vegetation type composition and canopy height distribution of the training data, and then presents the model results and comparisons with airborne LiDAR CHMs and existing canopy height products. The conclusion is presented in Section 5.

## 2. Data

### 2.1 Overview of the training datasets

The objective of this study is to generate CHM from remote sensing RGB imagery. Accordingly, the primary data types include remote sensing RGB images and spatially co-registered CHMs, or alternatively, airborne and UAV LiDAR point clouds or paired digital surface models (DSM) and digital elevation models (DEM) used for CHM generation.

The datasets were collected from multiple countries, including the United States, Australia, New Zealand, the Netherlands, Finland, Switzerland, Germany, China, and Brazil. Their spatial distribution is illustrated in Fig. 1, and detailed information is provided in Table 1.

For study areas where both airborne LiDAR point clouds and co-located airborne RGB imagery are available (*e.g.*, NEON sites in the United States and the Saihanba study area in China), The airborne RGB imagery was also incorporated into the training process to enhance sample diversity and representativeness.

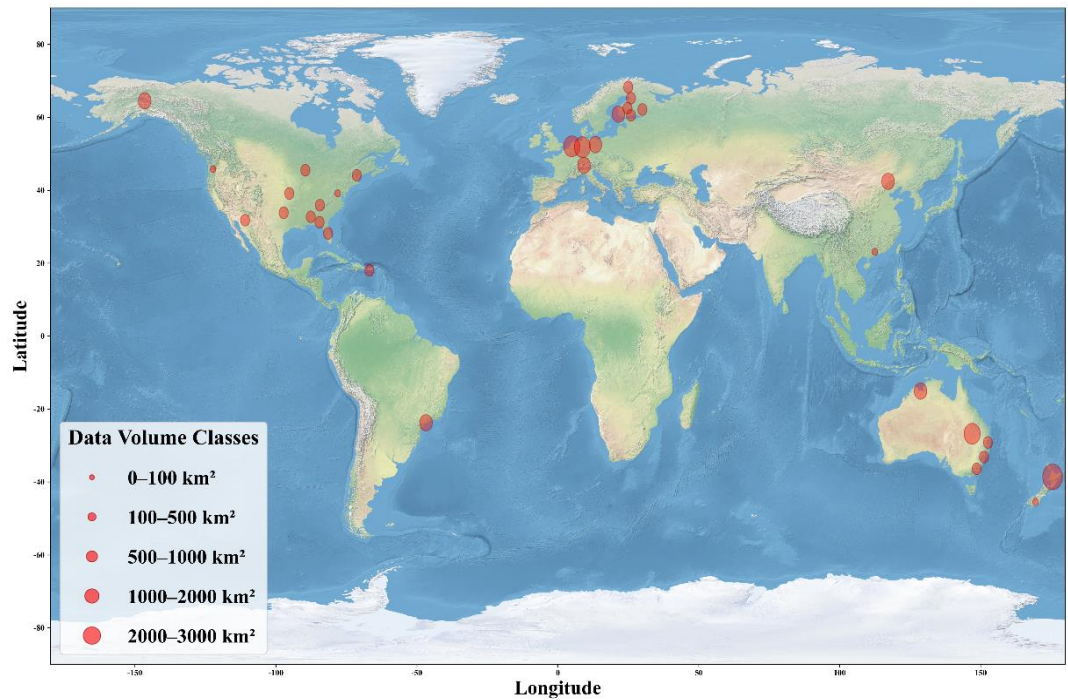

Fig. 1. Spatial distribution of the training sites. The base map is derived from Natural Earth

(https://www.naturalearthdata.com).

Table 1. Detailed information of the training sites.

| Country | Data types | Area (km$^2$) | Data acquisition year(s) |
|---|---|---|---|
| Australia | Point clouds, CHM | 2903 | 2016-2023 |
| United States | CHM | 2754 | 2017-2022 |
| New Zealand | DSM, DEM | 2326 | 2024 |
| Finland | Point clouds | 2196 | 2018 |
| Germany | DSM, DEM | 2181 | 2020-2023 |
| Netherlands | DSM, DEM | 1213 | 2023 |
| Brazil | DSM, DEM | 813 | 2017 |
| China | Point clouds | 678 | 2018-2022 |
| Switzerland | DSM, DEM | 673 | 2019 |

## 2.2 Satellite RGB imagery

Satellite RGB imagery was obtained from PlanetScope, using visually optimized 3 m spatial resolution RGB images. For each study area, a single image was selected from the summer season of the same year as the corresponding CHM acquisition (June 1 to August 30 for the Northern Hemisphere, and December 1 to February 28 for the Southern Hemisphere). The image selection criteria were as follows: cloud cover = 0, sun elevation > 50°, and view angle < 5°. If multiple images satisfied these criteria, the image with the smallest view angle was selected. If no image met all criteria, the constraints were relaxed to cloud cover < 10%, sun elevation > 60°, and view angle < 10°. If residual clouds were present in the selected image, cloud-covered regions were manually masked and excluded from both training and validation.

## 2.3 Airborne CHM generation

This section describes the procedures used to process different input data types (*e.g.*, LiDAR point clouds, paired DSM and DEM products) into CHMs with a unified UTM projection and a spatial resolution of 3 m.

For datasets provided as LiDAR point clouds, outliers were first removed, followed by an assessment of whether the point clouds had already been classified. For classified point clouds (*e.g.*, ground, vegetation, and non-vegetation classes such as water bodies and buildings), ground points and vegetation points were extracted to generate 3 m resolution DEMs and DSMs, respectively. Missing pixels in the DEM were filled using cubic spline interpolation, while missing pixels in the DSM (*e.g.*, over large water bodies or built-up areas) were assigned a value of −1. The CHM was then obtained as DSM − DEM, and negative canopy height values were set to 0. For unclassified point clouds, ground and non-ground points were separated using the cloth simulation filter (CSF) (Zhang et al. 2016b). DEMs and DSMs were subsequently generated, and CHMs containing heights of artificial structures (*e.g.*, buildings) were initially produced as DSM − DEM. To remove non-vegetation objects, vegetation and non-vegetation pixels were identified using RGB-based thresholding of the three channels or vegetation indices, *e.g.*, Normalized Difference Index and Excess Blue Index (Meyer and Neto 2008; Woebbecke et al. 1995). Pixels corresponding to non-vegetation areas with non-zero canopy height values were then set to 0, thereby excluding the influence of artificial structures on model training. For study areas with co-located airborne RGB imagery, this filtering procedure was based on the airborne RGB images. For areas without paired airborne RGB imagery, the same procedure was applied using PlanetScope imagery. Finally, all CHMs were reprojected to UTM.

For study areas where the initial data consisted of DSM and DEM products, CHMs containing artificial structures were first generated using DSM − DEM. Vegetation-only CHMs were then obtained by removing artificial structures using RGB imagery, followed by reprojection and resampling (using maximum value aggregation) to a UTM grid at 3 m resolution.

For datasets provided directly as CHMs, RGB imagery was used to determine whether the height of artificial structures had already been removed. If so, the CHMs were directly reprojected and resampled; otherwise, the height of artificial structures were removed using the same RGB-based filtering procedure prior to reprojection and resampling.

## 2.4 Airborne RGB imagery and land cover data

For study areas in the United States (National Ecological Observatory 2026b), China, the Netherlands, Germany, Finland, New Zealand, and Switzerland, airborne RGB imagery was acquired concurrently with airborne LiDAR data. These airborne RGB images were used both to assist in removing artificial structures from CHMs (Section 2.3) and, after being resampled to a 3 m resolution, to be directly paired with CHMs for model training.

Land cover information was obtained from GLC_FCS30D (Zhang et al. 2023), which classifies forests and shrublands into 13 detailed vegetation types and spans the period from 1985 to 2022. For training samples acquired after 2022, land cover types were assumed to remain unchanged in the absence of major natural or anthropogenic disturbances. This dataset was used to analyze the vegetation type composition of the training samples, with results presented in Section 4.1.

## 2.5 Comparison and validation datasets

Validation was conducted using datasets from the NEON SCBI site and four sites in China. The SCBI site covered an area of approximately 116 km², and its 1 m resolution CHM acquired in August 2021 was downloaded from the NEON data portal (National Ecological Observatory 2026a). The CHM was resampled to 3 m resolution using maximum value aggregation to match the resolution of the model-derived CHMs. For the four Chinese validation sites, airborne LiDAR point clouds were acquired between July and September 2024 and processed into 3 m resolution CHMs following the procedures described in Section 2.3.

In addition to LiDAR–derived CHMs, a 1 m resolution global CHM product generated by Tolan et al. (2024) was downloaded from Google Earth Engine and was hereafter referred to as MetaCHM. To ensure consistent spatial resolution, MetaCHM was also resampled to 3 m resolution using maximum value aggregation.

# 3. Methods

## 3.1 Related work on monocular depth estimation

Monocular Depth Estimation (MDE) aims to predict a dense depth map from a single RGB image, where depth is defined as the distance between each pixel and the imaging camera. Traditional depth estimation approaches, such as stereo matching and structure-from-motion (SfM), rely on multi-view geometric constraints and camera motion (Hartley and Zisserman 2003; Özyeşil et al. 2017). In contrast, MDE directly learns a depth mapping from a single image and is typically formulated as an end-to-end prediction problem driven by weakly supervised or unsupervised learning paradigms (Ming et al. 2021; Zhao et al. 2020).

Early MDE studies were predominantly based on CNN architectures, where prediction accuracy was improved through multi-scale feature extraction and deep feature fusion strategies (Eigen et al. 2014; Laina et al. 2016). Representative works included CNN-based depth regression networks that integrated deep continuous conditional random fields (CRFs) to refine depth predictions (Xu et al. 2017). These methods demonstrated the effectiveness of deep neural networks for MDE on benchmark datasets such as NYU Depth and KITTI.

With the emergence of Transformer architectures and large-scale datasets, MDE has gradually shifted from purely supervised learning toward paradigms that combine large-scale pretraining with weak or self-supervision. For example, ZoeDepth introduced a multimodal network architecture that jointly leveraged relative depth and metric depth supervision, improving both cross-scene generalization and metric scale accuracy (Bhat et al. 2023). To further enhance depth estimation performance on arbitrary images, Depth Anything proposed training a vision foundation model for MDE using large-scale unlabeled and synthetic data, combined with unsupervised and weakly supervised learning strategies (Yang et al. 2024a). The core idea of Depth Anything lay in constructing a massive unlabeled image dataset (approximately 62 million images) and incorporating carefully designed data augmentation and auxiliary semantic supervision, enabling the model to inherit rich semantic priors and achieve strong generalization capability (Yang et al. 2024a). Building upon this framework, Depth Anything V2 further improved both model capacity and practical performance. Compared with its predecessor, Depth Anything V2 replaced a portion of real annotated data with a large amount of high-quality synthetic images, effectively reducing errors induced by noisy labels. In addition, a higher-capacity teacher model was trained on synthetic data, and the large-scale pseudo labels generated by the teacher were used to supervise student models, leading to improved fine-scale detail representation and enhanced robustness across diverse scenes (Yang et al. 2024b).

**3.2 Training Depth Anything V2 with paired RGB images and CHMs**

Depth Anything V2 was trained on large-scale high-quality synthetic depth data within a teacher–student learning framework, as described in its original formulation. The pretrained model demonstrated strong zero-shot generalization ability across multiple public benchmarks, including KITTI, NYU Depth V2, ETH3D, ScanNet, and DIODE. Since CHM represented the vertical distance between the canopy surface and the ground, while the pretrained Depth Anything V2 model predicted the distance between image pixels and the camera, a representation mismatch existed between the two depth definitions. To align the CHM with the semantic meaning of depth in the pretrained model, we assumed a fixed maximum canopy height of 50 m and converted CHM values into pseudo-depth by subtracting the CHM from this maximum height. This transformation preserved the relative depth ordering and spatial structure while enabling effective utilization of the pretrained weights.

The ViT-L backbone was selected as the model architecture for training in this study. The main training parameters were set as follows: 25 epochs, a batch size of 4, a learning rate of $5 \times 10^{-6}$, a maximum depth of 500, and a minimum depth of 0. Model training was conducted on a server equipped with two NVIDIA L20 GPUs (48 GB memory each) and an Intel® Xeon® Platinum 8457C CPU. The trained model was hereafter referred to as DepthV2CHM.

## 3.3 The metrics used for comparison and analysis

In this study, three quantitative metrics—Bias, Mean Absolute Error (MAE), and Root Mean Square Error (RMSE)—were used to evaluate the accuracy of the estimated CHMs. Bias characterized the systematic deviation of model estimates relative to the reference values, indicating whether the model tended to overestimate (positive values) or underestimate (negative values) canopy height. Bias was computed as:

$$Bias = \frac{1}{N}\sum_{i=1}^{N}(\hat{y}_i - y_i) \quad (1)$$

where $\hat{y}_i$ denoted the value of the $i$th pixel in the Depth2CHM or MetaCHM, $y_i$ denoted the corresponding pixel value in the airborne LiDAR–derived CHM, and $N$ was the total number of pixels.

The MAE measured the average magnitude of the absolute differences between the estimated and reference values, reflecting the overall error level while being less sensitive to outliers. It was calculated as:

$$MAE = \frac{1}{N}\sum_{i=1}^{N}|\hat{y}_i - y_i| \quad (2)$$

The RMSE evaluated the overall magnitude and dispersion of estimation errors. By assigning greater weight to larger errors, RMSE was more sensitive to outliers and was defined as:

$$RMSE = \sqrt{\frac{1}{N}\sum_{i=1}^{N}(\hat{y}_i - y_i)^2} \quad (3)$$

Additionally, the Structural Similarity Index (SSIM) was employed to assess the spatial consistency between the estimated CHM (*i.e.*, Depth2CHM and MetaCHM) and the airborne LiDAR–derived CHM. In this study, the sliding window size for SSIM computation was set to 11 × 11 pixels, corresponding to 33 m × 33 m. The SSIM for the $i$th window was computed as:

$$SSIM_i = \frac{(2\mu_{\hat{y}_i}\mu_{y_i} + C_1)(2\sigma_{\hat{y}_i,y_i} + C_2)}{(\mu_{\hat{y}_i}^2 + \mu_{y_i}^2 + C_1)(\sigma_{\hat{y}_i}^2 + \sigma_{y_i}^2 + C_2)} \quad (4)$$

where $\mu_{\hat{y}_i}$ and $\mu_{y_i}$ were the mean values of the estimated CHM and the reference airborne CHM within the $i$th window, respectively; $\sigma_{\hat{y}_i}^2$ and $\sigma_{y_i}^2$ were the corresponding variances; and $\sigma_{\hat{y}_i,y_i}$ was the covariance between the two images within the window. The constants $C_1$ and $C_2$ were defined as:

$$C_1 = (K_1 L)^2 \tag{5}$$

$$C_2 = (K_2 L)^2 \tag{6}$$

where $K_1 = 0.01$ and $K_2 = 0.03$, and $L$ represented the dynamic range of the image pixel values. The overall SSIM for an image was computed as the mean SSIM across all sliding windows. SSIM values ranged from −1 to 1, with higher values indicating greater similarity in spatial structure between the two images.

## 4. Results and analysis

### 4.1 Vegetation type composition and canopy height distribution of the training dataset

Except for the closed mixed leaf forest type, the training dataset covered nearly all forest and shrub vegetation types defined in GLC_FCS30D (Fig. 2a), indicating that the training data included a wide range of vegetation types with strong representativeness. The five most dominant vegetation types were closed evergreen needle-leaved forest, closed deciduous broadleaved forest, closed evergreen broadleaved forest, open deciduous needle-leaved forest, and shrubland. Broadleaved forest types accounted for 38.48% of the training samples, while needle-leaved forest types accounted for 46.07%. Mixed forest types represented only 0.84%, and shrubland accounted for 14.60% of the total samples. The distribution of CHM values greater than 0 m in the training dataset is shown in Fig. 2(b). Most CHM values were concentrated between 0 and 25 m, accounting for 90.03% of the samples. Such diversity in vegetation types and canopy height ranges was expected to reduce vegetation-type bias during training and improved the robustness of the model across different forest conditions.

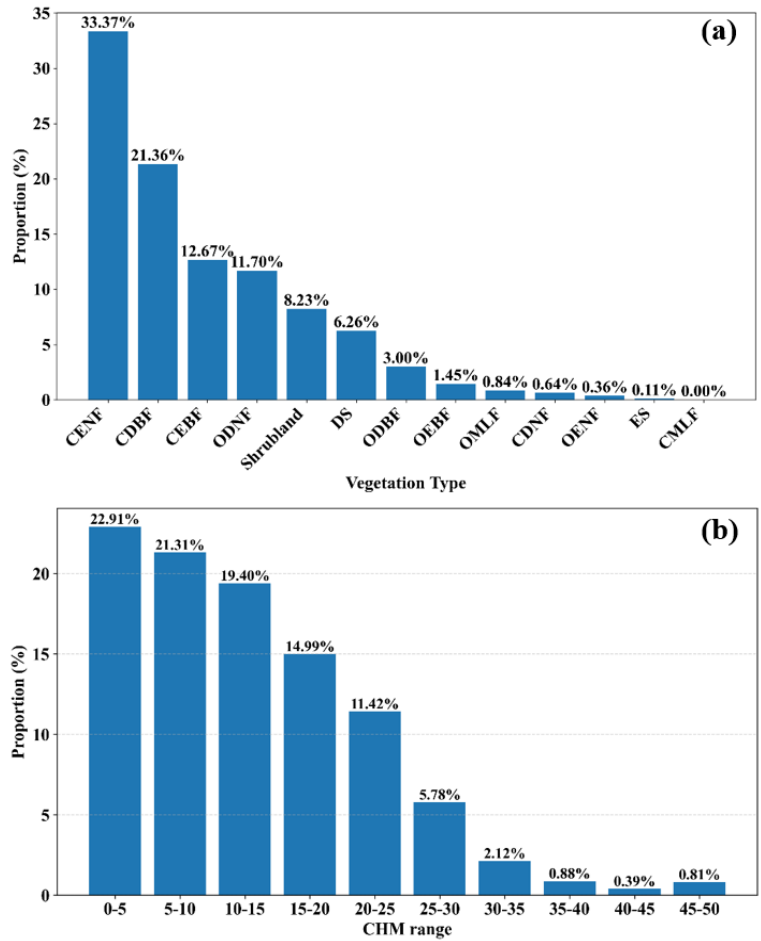

Fig. 2. Distribution of vegetation types and canopy height values (> 0 m) in the training dataset. (a) Distribution of forest and shrub vegetation types. The x-axis shows abbreviations of vegetation types, where C and O denote closed and open, E and D denote evergreen and deciduous, B and N denote broadleaved and needle-leaved, and F and S denote forest and shrubland, respectively. For example, CENF represents closed evergreen needle-leaved forest. (b) Distribution of CHM values greater than 0 m in the training dataset.

### 4.2 Accuracy assessment based on LiDAR–derived CHMs

This section evaluated the performance of Depth2CHM and MetaCHM using airborne CHM data acquired in 2021 at the SCBI site in the United States and in 2024 at four validation sites in China. It should be noted that none of these areas were included in the training process of this study. The Depth2CHM estimates were derived from PlanetScope imagery acquired in the same year as the corresponding airborne CHM data, whereas MetaCHM was generated using Maxar Vivid2 mosaic imagery collected during 2017–2020.

**4.2.1 Results at the SCBI validation site in the United States**

The spatial distributions of the three CHM products are shown in Fig. 3. Clear differences in spatial patterns could be observed between MetaCHM and the airborne CHM. MetaCHM exhibited an overall underestimation, with a SSIM of 0.43. In contrast, the spatial distribution of Depth2CHM was more consistent with that of the ALS CHM, without evident overall underestimation or overestimation, resulting in a higher SSIM value of 0.56.

Furthermore, Figs. 4 and 5 present the scatter plots and difference distribution histograms of MetaCHM and Depth2CHM against the airborne CHM, respectively. As shown in Fig. 4(a), the scatter points of MetaCHM were predominantly distributed below the 1:1 line, indicating a pronounced underestimation, with a bias of −4.30 m, an MAE of 5.58 m, and an RMSE of 7.50 m. By contrast, the scatter points of Depth2CHM were more tightly clustered around the 1:1 line (Fig. 4b), with a bias of 0.41 m, an MAE of 4.24 m, and an RMSE of 5.75 m. In addition, the difference distribution of Depth2CHM was more concentrated within the range of −5 m to 5 m compared with MetaCHM (Fig. 5). Specifically, 66.40% of the pixels of Depth2CHM fell within this error range, whereas the corresponding proportion for MetaCHM was 53.94%. Overall, these results indicated that Depth2CHM achieved higher accuracy than MetaCHM at the SCBI site.

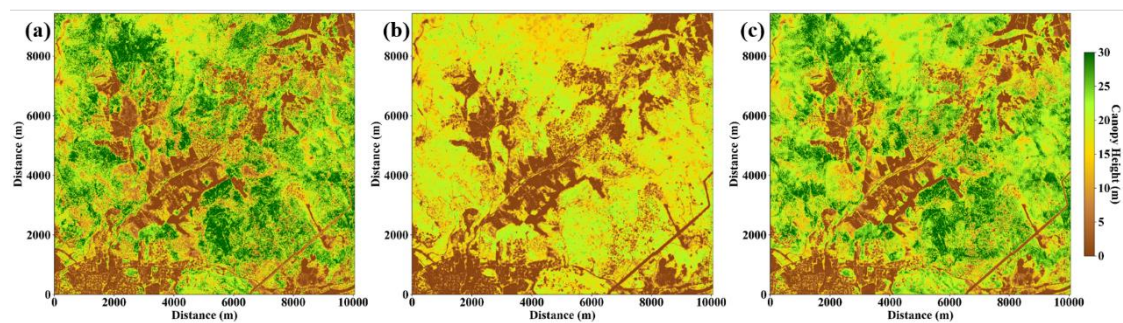

Fig. 3. Spatial distribution of canopy height at the SCBI site derived from (a) airborne LiDAR point clouds, (b) MetaCHM, and (c) Depth2CHM.

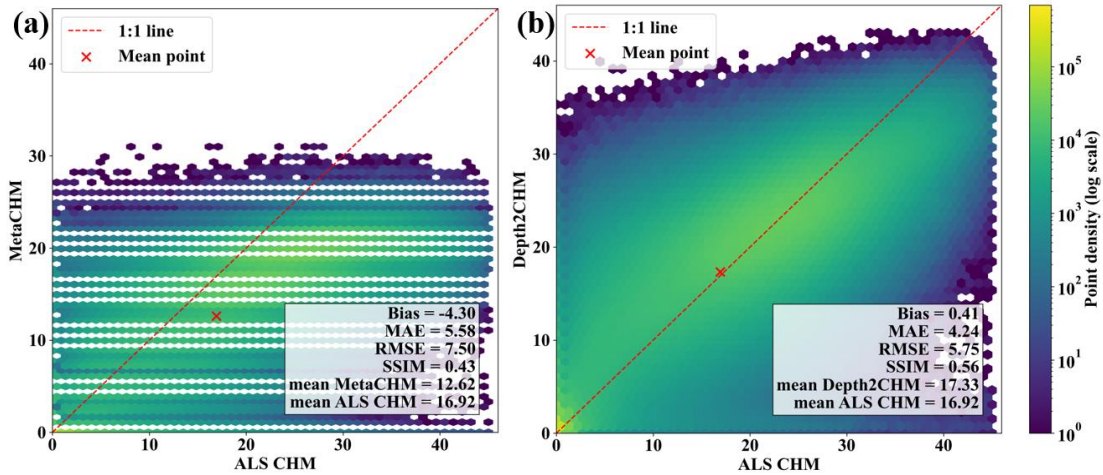

Fig. 4. Scatter plots of (a) MetaCHM and (b) Depth2CHM versus the airborne LiDAR–derived CHM at the SCBI site. Because canopy heights in the MetaCHM product are reported as integer values, horizontal gaps parallel to the x-axis appear where certain height values are absent.

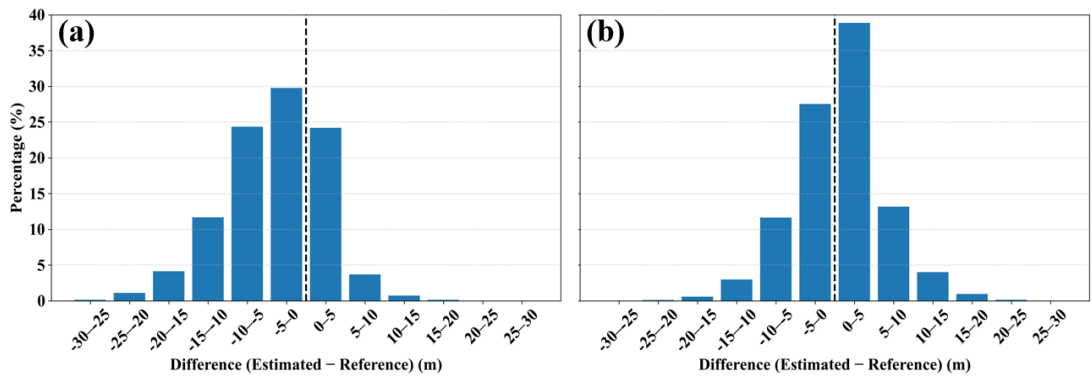

Fig. 5. Distribution of differences between (a) MetaCHM and the airborne LiDAR–derived CHM, and between (b) Depth2CHM and the airborne LiDAR–derived CHM at the SCBI site.

### 4.2.2 Results at the Chinese validation sites

The spatial distributions of the three CHM products over the four validation sites in China are shown in Fig. 6. Similar to the results observed at the SCBI site in the United States, MetaCHM (Figs. 6e–6h) exhibited an overall underestimation compared with the airborne LiDAR -derived CHM (Figs. 6a–6d), with pronounced differences in the spatial distribution of canopy height values. The corresponding SSIM between MetaCHM and the airborne CHM was relatively low (0.32). In contrast, the spatial distribution of Depth2CHM was much more consistent with that of the airborne CHM (Figs. 6i–6l), yielding a substantially higher SSIM value of 0.69.

The scatter plots and difference distribution histograms of the three CHM products are presented in Figs. 7 and 8, respectively. The scatter points of MetaCHM were predominantly located below the 1:1 line (Fig. 7a), and pixels with negative errors accounted for as much as 92.86% of the total, indicating a pronounced underestimation. Quantitatively, MetaCHM showed a bias of −2.99 m, an MAE of 3.25 m, and an RMSE of 4.65 m. By comparison, the scatter points of Depth2CHM were more tightly clustered around the 1:1 line (Fig. 7b). The proportions of pixels with negative and positive errors were 47.51% and 52.49%, respectively, indicating no systematic underestimation or overestimation. Moreover, the proportion of pixels with absolute errors smaller than 3 m reached 79.02% for Depth2CHM, which was substantially higher than that of MetaCHM (55.92%). Correspondingly, Depth2CHM achieved a bias of 0.59 m, an MAE of 1.64 m, and an RMSE of 2.54 m. Overall, these results demonstrated that Depth2CHM attained higher accuracy than MetaCHM across the Chinese validation sites.

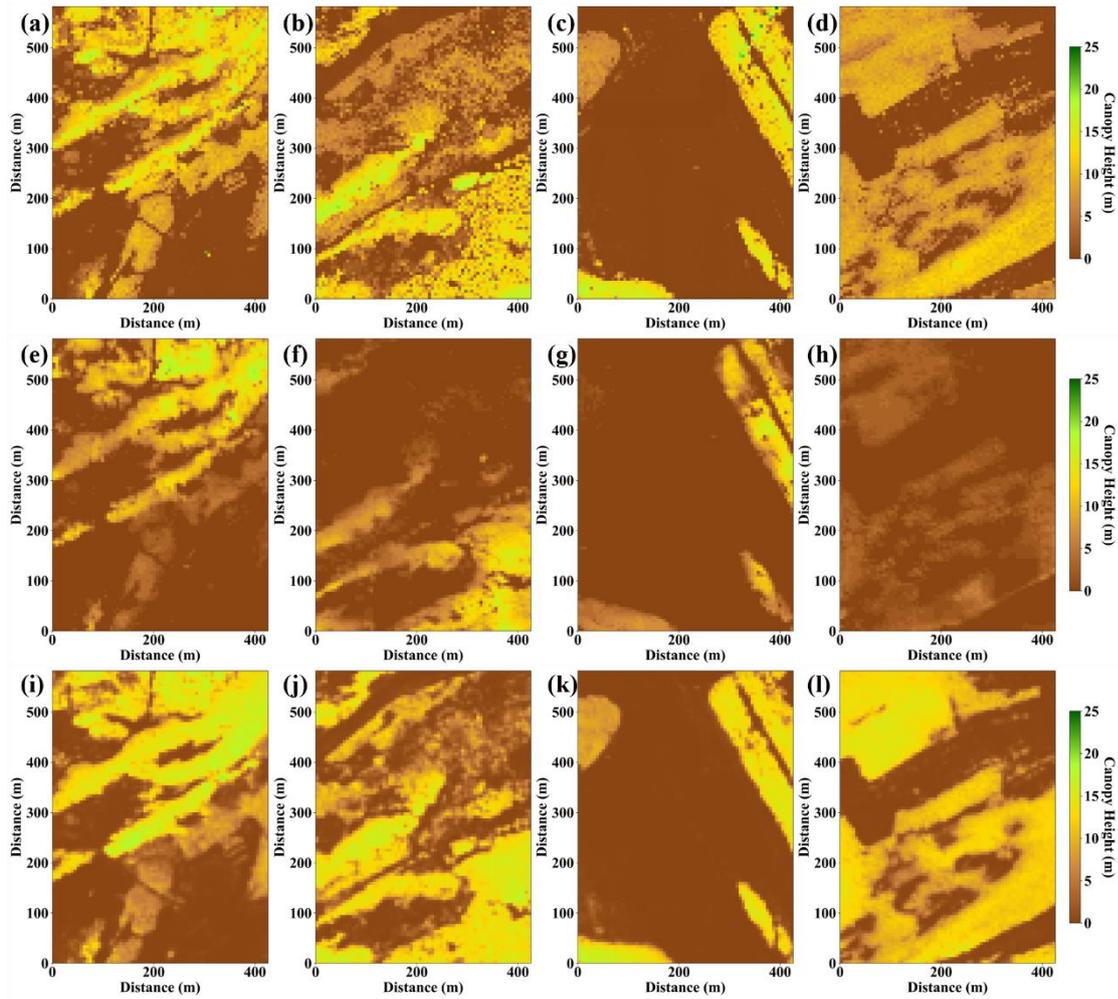

Fig. 6. Comparison of canopy height spatial distributions derived from (a)-(d) airborne LiDAR point clouds, (e)-(h) MetaCHM, and (i)-(l) Depth2CHM across four validation sites in China.

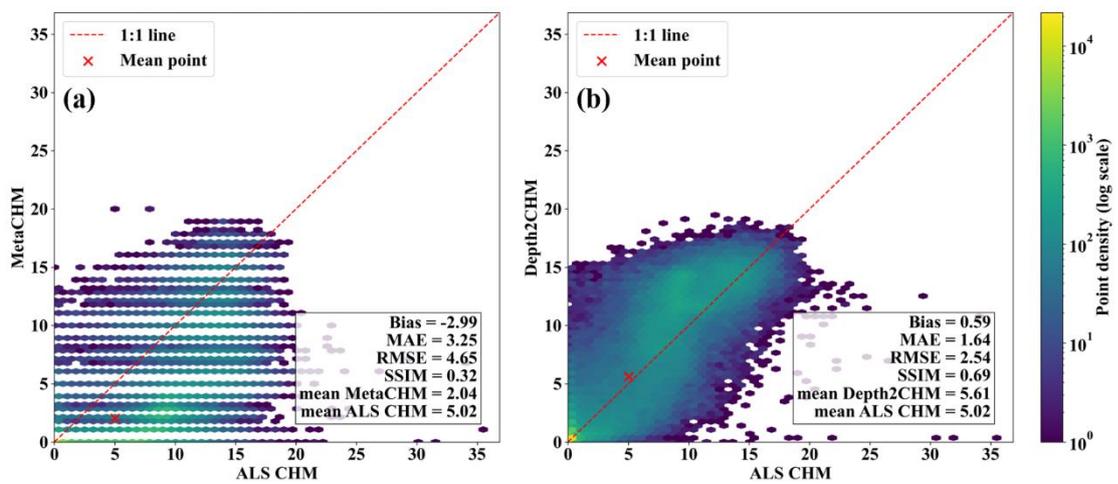

Fig. 7. Scatter plots of (a) MetaCHM and (b) Depth2CHM versus the airborne LiDAR–derived CHM across four validation sites in China. Because canopy heights in the MetaCHM product are reported as integer values,

horizontal gaps parallel to the x-axis appear where certain height values are absent.

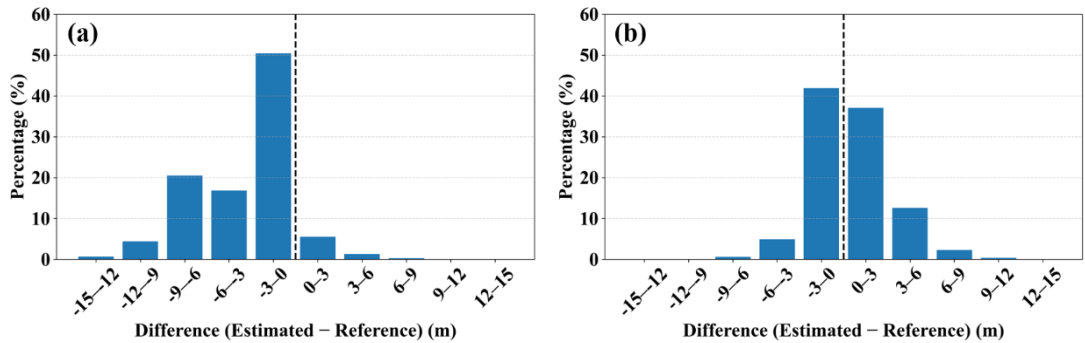

Fig. 8. Distribution of differences between (a) MetaCHM and the airborne LiDAR–derived CHM, and between (b) Depth2CHM and the airborne LiDAR–derived CHM across four validation sites in China.

## 5. Conclusion

Large-scale, spatially continuous, high-resolution canopy height mapping is essential for studies of carbon and water cycles. In this study, we trained a state-of-the-art monocular depth estimation model, Depth Anything V2, using globally distributed airborne LiDAR–derived canopy height models (CHMs) covering diverse vegetation types and a total area of approximately 16,000 km², together with co-registered 3 m resolution PlanetScope RGB imagery.

The trained model, referred to as Depth2CHM, was capable of directly producing 3 m resolution CHM estimates from PlanetScope RGB imagery alone. Validation using independent datasets demonstrated that Depth2CHM could accurately estimate canopy heights across a wide range of forest structures, with a mean bias of approximately 0.5 m. Compared with an existing 1 m resolution global canopy height product, Depth2CHM reduced the mean absolute error by approximately 1.5 m and the root mean square error by approximately 2 m.

## Acknowledgements

This study made use of multiple open-access geospatial datasets provided by national and regional mapping agencies. Data from the National Ecological Observatory Network were used in this research. The National Ecological Observatory Network is sponsored by the United States National Science Foundation and operated by Battelle. High-resolution orthorectified camera imagery mosaic data and ecosystem structure data were obtained through the National Ecological Observatory Network Data Portal.

Elevation and surface model data were additionally obtained from several governmental open data platforms, including the Actueel Hoogtebestand Nederland provided via the Publieke Dienstverlening Op de Kaart services in the Netherlands, the SwissSURFACE3D and SwissALTI3D height models provided by the Federal Office of Topography swisstopo in Switzerland, laser scanning data provided by the National Land Survey of Finland, elevation datasets provided by the Geoportal North Rhine-Westphalia and the Berlin Geodata Infrastructure in Germany, elevation data provided by the GeoSampa platform of the Municipality of São Paulo in Brazil, raster elevation data provided by Land Information New Zealand, and elevation datasets provided by the Foundation Spatial Data Framework Elevation service in Australia.

All datasets are provided as open geospatial data by the respective agencies and were accessed in accordance with their data use and attribution guidelines. We gratefully acknowledge these organizations for making their valuable data publicly available, which has been essential for the success of this research.